# Computational Intelligence Characterization Method of Semiconductor Device

Eric Liau
Infineon Technology AG
Design Analysis and Verification
Balanstraße 73 D-81541
Munich Germany
Email: eric.liau@infineon.com

Doris Schmitt-Landsiedel
Lehrstuhl für Technische Elektronik
Technische Universität München
Theresienstr. 90, Gebäude N3
80290 Munich Germany
Email: dsl@ei.tum.de

**Abstract**

*Characterization of semiconductor devices is used to gather as much data about the device as possible to determine weaknesses in design or trends in the manufacturing process. In this paper, we propose a novel multiple trip point characterization concept to overcome the constraint of single trip point concept in device characterization phase. In addition, we use computational intelligence techniques (e.g. neural network, fuzzy and genetic algorithm) to further manipulate these sets of multiple trip point values and tests based on semiconductor test equipments, Our experimental results demonstrate an excellent design parameter variation analysis in device characterization phase, as well as detection of a set of worst case tests that can provoke the worst case variation, while traditional approach was not capable of detecting them.*

## 1. Introduction

There is a distinction between semiconductor production verification tests and engineering characterization and analysis tests. Production testing determines if the device meets its design specification and, if it does not, stops testing on first fail, bins the device and goes on to the next device. In contrast, the methodology for characterization is a kind of closed loop test; that is, a test repeated many times within a specific timing edge varied with a range, looking for the pass/fail point of an associated parameter, and this is called trip point as shown in figure 1. The key to this process is discovering the trip point as accurately as possible, and determines the exact limits of device operating values. It is important to note that device specifications and limits of operating values are fixed in the design phase (such as clock frequency), and are not expected to vary during device operation phase. In reality, the exact limits of device operating values can be better or worse than the expected design specification due to different input tests (pattern and test condition variation), and semiconductor process variation. Ideally, we want characterization tests for the worst case, because it is easier to evaluate than average cases and devices passing this test will work for any other conditions. This is done by selecting a pre-defined test that results in a chip pass/fail decision. Then select a statistically significant sample of devices, and repeat the test for every combination of two or more environmental variables. This essentially means repetitively applying functional tests and measuring the limits of various DC or AC parameters, such as supply voltage or clock frequency. This set of information helps to define the final device specification at the end of the characterization phase, and develop a production test program in manufacturing test.

Various methods for characterization and altering test values to find a device's trip point, such as successive approximation, binary search, and linear search are used in ATE [1:7] [16] today. A linear search starts at one boundary and steps through a specified resolution until the stage changes or the end boundary is reached. The trip point is a device pass. This approach has some disadvantages. If the resolution is small, the search can be time consuming. If the specification parameter changes over time due to device heating or other factors, an inaccurate reading could result. If the device is failing (or passing) over a large range of values the entire search must be run for several different ranges to reach this conclusion. A binary search method uses a divide-by-two approach. The delta between the last known true and last known false condition are halved until the trip point is found. The search switches directions (toward the starting point and ending point) every time the condition changes from pass to fail or from fail to pass. The successive approximation searches between two values, using one of the boundary values and a value half way in between. If both produce the same results, the search continues to the other end of boundary. If each produces difference results, then the same search continues between the passing and failing points until the trip point is found. This approach is similar to binary search, except the successive approximation uses an algorithm that can sense a drifting specification parameter and make a judgment as to the direction and span of the search. This method is recommended for device performance characterization at most of the ATE [1:7] today. However, all of these characterization approaches can not address the problem of potential specification variation due to different tests. This is because the exact limits of device operating performance are heavily dependent on input tests. Input tests are referred to input test patterns and test conditions. Thus, a set of pre-defined tests with a single trip point analysis can not guarantee that the trip point stays within the specification under all admissible conditions. It is practically impossible to determine the true worst case test manually using a deterministic method. This finally leads to the major technical challenges: How to select a set of worst case tests that can provoke the worst case variation against specification? How can we automate this process intelligently? This paper solves the problem efficiently using computational intelligence techniques with industrial ATE.

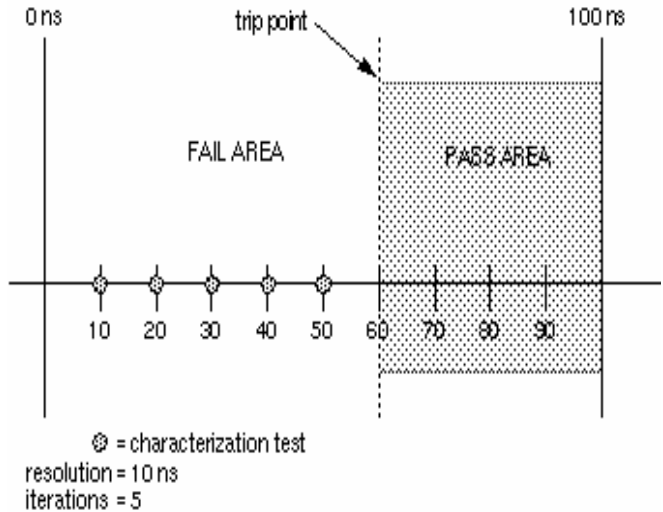

Fig. 1. Single trip point concept in device characterization process

## 2. Contribution

Comparing to the traditional device characterization concepts [1-7] [15-16], our work has the following contributions [11]:

- We propose multiple characterization trip point concept instead of conventional single trip point method.
- We develop a search method: search until trip point technique, to reduce the repetition of measurement during characterization phase. This method ultimately speeds up the searching time of worst case test in characterization process.
- We use neural network (NN) to learn from a set of input tests and their corresponding characterization trip points via ATE. In addition, we propose to use fuzzy set theory to encode the characterization trip point information. In operation phase, neural network will perform a classification task to identify the worst case test. Finally, this set of pre-selected worst case tests will be further optimized by genetic algorithm (GA) based on the fitness of the trip point value obtained from the ATE. Final set of worst case tests can be re-simulated or analyzed in detail with ATE (e.g. wafer probing analysis) to localize the design weakness efficiently.

The rest of the paper is organized as follows. In section III, we describe the novel multiple trip point characterization concept. In section IV, we describe the search until trip point technique to improve the time efficiency during the search of worst case test in characterization process. In section V, we describe the implementation of computational intelligence techniques for device characterization with industrial ATE. The experimental results are presented in Section VI. Section VII concludes the paper.

## 3. Multiple Characterization Trip Point

There are many cases where several design parameters were measured and passed within the design specification using conventional approaches, but this method failed to detect the worst case behavior in real application. The major weakness is that it relies only on a small set of pre-defined deterministic tests (patterns and test conditions), and a single trip point measurement. Typical technical efforts of semiconductor device characterization are only focused on how to get this trip point as accurate as possible. However, to guarantee that the device meets the design specification under all test conditions, we can not rely on a few single tests and single trip points. Trip point values could fluctuate easily with respect to different non-deterministic random tests, such as bus control signals in real application board. Therefore, we propose to determine the worst case trip point value based on a set of multiple trip point measurements with respect to different non-deterministic random tests.

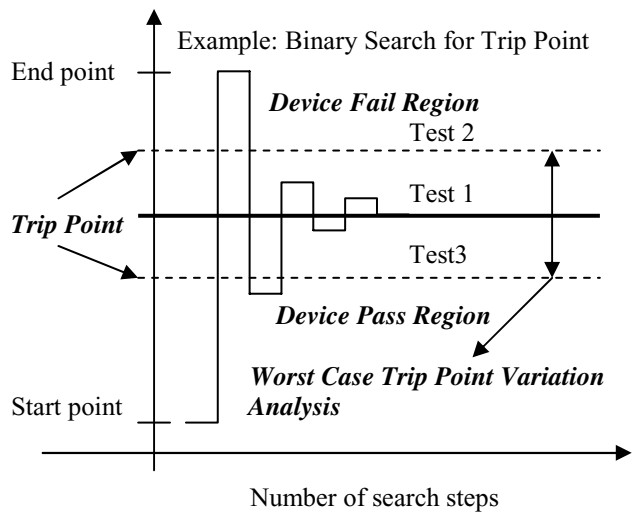

Fig. 2. Multiple trip point concept in device characterization process

For the procedure in figure 2, we use the random test generator based on [9-10], combined with a device characterization algorithm such as binary search or successive approximation. In order to pin-point the potential worst case test sequences more precisely, we define small test sequences in between 100 to 1000 vector cycles for each characterization measurement of a single trip point. As a result, we obtain a set of design specifications DSV which is equal to the trip point values obtained from different input tests $T_n$ where each test (e.g. test 1, 2 and 3 in figure 2) produces individual trip point values. (*N* is number of tests.)

$$DSV = TPV(T_1,...,T_N) \quad (1)$$

## 4. Search Until Trip Point Algorithm

There are several technical constraints about device characterization testing to keep in mind. First, characterization is a lengthy process since it involves multiple repetitions of a test. Second, the search algorithm requires that starting points be chosen on both sides of the good to bad crossover as shown in figure 2. It is easy to underestimate the range required and to choose starting points that are both good. Very generous starting ranges should be selected. Third, characterization tests are aimed at characterizing independent parameters one at a time. The test conditions must be such that only the parameters being tested can cause test failure. All the other parameters must be relaxed so they can not cause test failures and false convergence.

The above constraints create a major technical issue of measurement speed if a multiple trip point concept is used. To solve this problem practically, we propose to use "search until trip point algorithm" during multiple trip point characterization process, as in figure 3: assume T is a set of random input tests, and the upper boundary value *P* of the pass region is smaller than the lower boundary *F* of the fail region, such as specified operating frequency of the device is 100MHz and the device will fail if operating frequency is further increased above 110MHz. In order to have a generous starting range, we defined the starting frequency is *S1*=80MHz, and ending frequency is *S2*=130MHz. So the characterization range is *CR*=50MHz and we use successive approximation to perform the first trip point detection using any initial random test as in equation (2) based on equation (1). The first trip point value is then defined as *RTP*=reference trip point. And the next search will be based only on *RTP* conditions instead of full characterization range *CR*. If the second test passed (*N>1*) with *RTP*, then *RTP+SF* is used, where *SF* is defined as search factor resolution, and it is a programmable variable such as 1MHz or 2MHz per step and *IT* is defined as search iterations; thus *SF(IT)* can be defined as $SF(1) \times IT$. SF will further increase with *IT* until the device shows the first failure result and the trip point is detected. On the other hand, if the second test failed with *RTP,* then *RTP-SF(IT)* is used. *SF(IT)* is further decreased until the device shows the first working result again and the trip point is discovered. Instead of equation (3), we use equation (4) if the specification value of *P*=pass region is greater than the *F*=fail region. The major motivation of using equation (3) and (4) for multiple trip point method is that the variations of semiconductor device parameters or performance values are only expected in a very narrow range with respect to different input tests if the devices are properly designed. Therefore, it is not necessary to search through the whole "generous range" for multiple repetitions of trip point measurement that would cause a very lengthy process, since *CR(IT)* is much larger than *SF(IT)* as shown in figure 3. In addition, In case of unexpected drift of design performance vs target specification due to unexpected design weaknesses provoked by a set of worst case tests, our proposal is flexible enough to detect the drift while keeping smallest effort of searching for the trip point value based on *RTP*. This ultimately leads to huge savings of measurement time and guaranteed automatic convergence, keeping the test time as low as possible.

$$RTP = TPV(T_N)_{N=1} \quad (2)$$

$$TPV(T_N)_{N>1} = \begin{cases} (RTP + SF(IT)) & P<F \quad Pass \\ (RTP - SF(IT)) & P<F \quad Fail \end{cases} \quad (3)$$

$$TPV(T_N)_{N>1} = \begin{cases} (RTP + SF(IT)) & P>F \quad Fail \\ (RTP - SF(IT)) & P>F \quad Pass \end{cases} \quad (4)$$

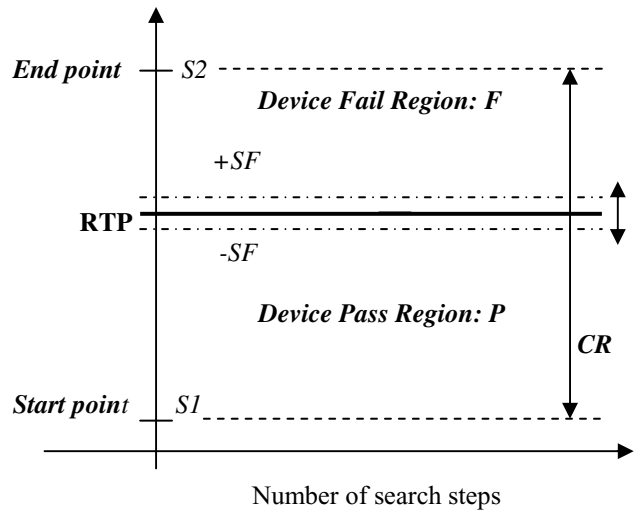

Fig. 3. Formulation of Search until Trip Point Algorithm

## 5. Intelligent Device Characterization

Today, what is missing in typical device characterization concepts with industrial ATE is that the test system is not designed to perform the worst case device characterization. Instead ATE is used to detect the trip point as accurate as possible based on a set of pre-defined patterns. A pre-defined test is based on deterministic way of testing the circuit. It does not for sure emulate the worst case application condition, and this ultimately leads to potential application failures, even if the circuit has passed all deterministic characterization tests. On the other hand, it would be a huge work if we try to analyze all different combinations of test sequences and specifications. To solve this limitation, we change the major objective of device characterization, focusing only on how to accurately detect the worst case test that can provoke the worst case performance vs. specification variation, while keeping the time of measurement as low as possible using the techniques proposed in section 2 and 3. In addition, we combine computational intelligence techniques with industrial ATE to perform learning of device characterization and the worst case test classification task. To implement this concept, we re-configure our previous work [9][10] to use it in semiconductor device characterization. The completed device characterization learning and optimization scheme can be described as follows in figures 4 and 5.

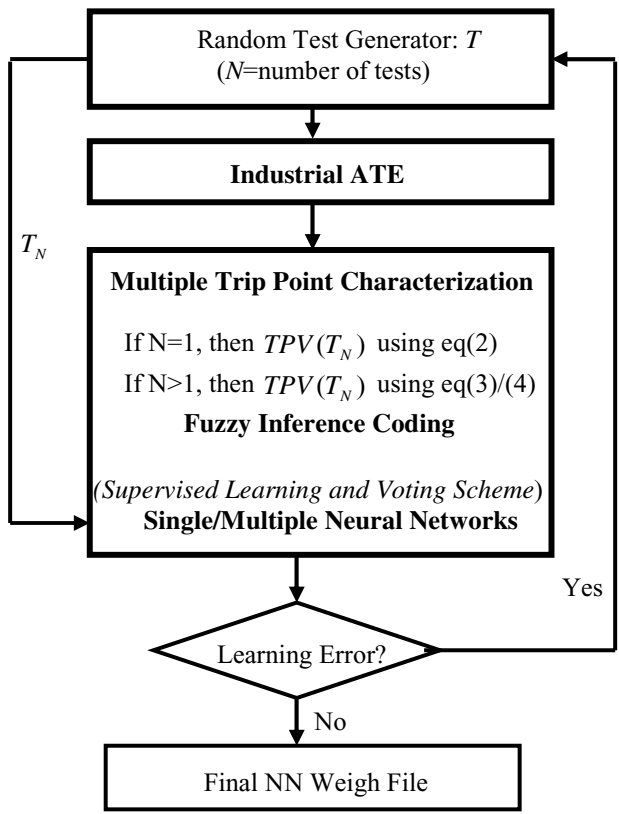

Fig. 4. Intelligence Device Characterization Learning Scheme with industrial ATE

- (1) To measure how confident the neural net is in its classification, we propose to use the NN voting machine algorithm, such that multiple NNs are trained on different subsets of the training input tests, then vote in parallel on unknown input tests. Thus, the first step is presenting a random test to ATE and neural network modules continuously.
- (2) Detect the first reference trip point *RTP* using equation (2), and search for the subsequent trip point using equation (3) or (4) depending on the search parameter conditions.
- (3) Trip point value coding using either fuzzy set data [8] or simple numerical coding; then NN starts to learn from input random tests and supervised by ATE detects *TPV* value from (2).
- (4) The confidence in the classification is determined by averaging the mean error for each network (i.e. consistency check). After that, NN will continue learning with iterative network learnability and generalization check [12-14] until learning and generalization error is small enough; otherwise go back to (1).
- (5) At the end of NN learning, a NN weight file is generated. This file will be used in classification task of worst case test based on only software computation without measurement in optimization phase as in figure 5.

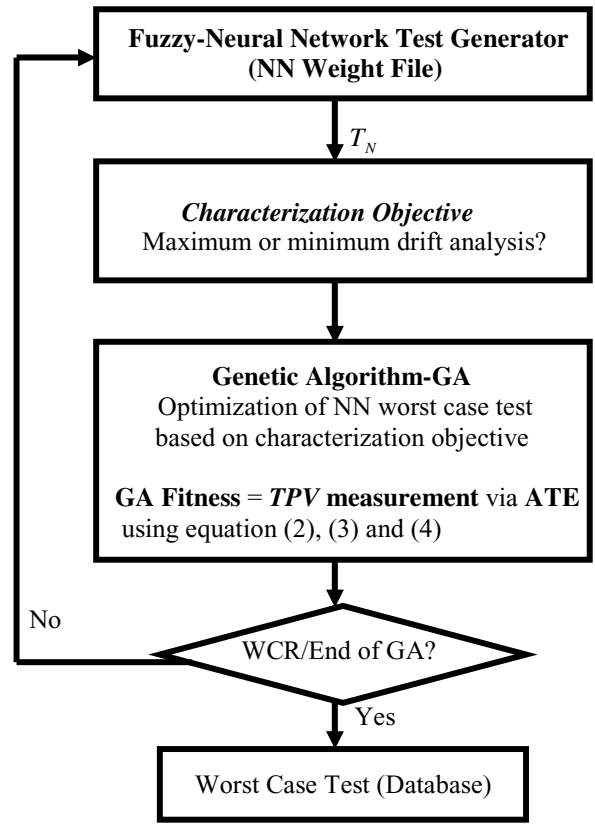

Fig. 5. Intelligence Device Characterization Optimization Scheme with industrial ATE

- (1) A number of GA test populations are initialized by a set of sub-optimal tests selected by fuzzy-neural network test generator based on its previous learning experience (NN weight file).
- (2) Define the characterization objective: generating a worst case test that can provoke the worst case characterization parameter drift, such

- as drift to the maximum value, or drift to the minimum value.
- (3) Test optimization starts using GA [12-14] based on its characterization objectives (2). The GA fitness=*TRV* evaluation is based on equation (2), (3) and (4) depending on $T_N$ conditions as described in section 3 and 4.
- (4) GA optimization process continues until GA fitness value can not improve anymore. Then go to (1) and a brand new population will start GA again. This process will continue until either it reaches the maximum optimization steps or the worst case is detected based on worst case ratio theorem. At last, final worst case tests are generated and stored in the database.

To classify the tests in the GA optimization process, we use the Worst Case Ratio (WCR), given in equation (5) and (6) for a parameter value $v_a$ obtained by ATE analysis in test number n, when N tests are performed in the total analysis time:

$$WCR(N) = \text{Max } |v_a(n)/v_{max}|, \quad 0<n\leq N \quad (5)$$

$$WCR(N) = \text{Min } |v_{min}/v_a(n)|, \quad 0<n\leq N \quad (6)$$

Where, $v_{max}$ or $v_{min}$ are specified max or min values, respectively, for parameters such as voltage, current, or timing deviation. GA classifications (see figure 6) could then be

Pass        $0 \leq WCR \leq 0.8$

Weakness    $0.8 < WCR \leq 1$

Fail        $WCR > 1$

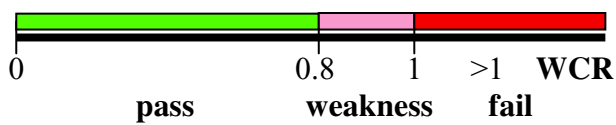

Fig. 6. Definition of Worst-Case Ratio WCR

The worst case tests are given by the largest values of WCR. In order to deal with two different types of chromosomes – test sequences and test conditions - we have developed a GA method evolving multiple populations of different individuals over a number of generations. A fitness value is assigned to each individual in the GA population. According to the analysis task, the fitness can be power consumption, peak current, voltage or other functionalities obtained from ATE. At the end of the complete iterative analysis, a final set of worst case tests is identified, covering all considered fitness variables. Functional failure patterns (if any) are stored separately. Then, we further analyze the potential design weaknesses and functional failures, using a transistor-level simulator and/or ATE.

Finally, a major achievement of our approach is the way of coding test inputs and measurement values from ATE. It is very complicated to model a NN with multiple output classification ability. Thus we propose to pre-select a set of DC or AC critical parameters; and generate NNs individually for each parameter or each characterization analysis task. Finally, based on our prior characterization experiences, we strongly recommend to use fuzzy variables to encode measurement values as fuzzy logic can describe more than one analysis parameter; such as if A and B and C, then D is *quite close* to the limit of the target device-spec.

## 6. Experimental Results

We have tested our approach using 140nm technology memory test chip. In our experiment, we select data output valid time $T_{DQ}$ (spec = 20ns) to perform the worst case parameter variation analysis. The data output valid time is defined as data valid time with respect to address changes. The smaller the $T_{DQ}$ value, the longer the required data valid time as show by the arrow direction in figure 7. Thus, the minimum value is the worst case, since the processor will have to wait for a longer time to read the valid information from the memory chip. Obviously, $T_{DQ}$ is test dependent, and the objective of our experiment is to detect a set of worst case tests that can provoke the worst case minimum $T_{DQ}$ value using equation (6)-minimization.

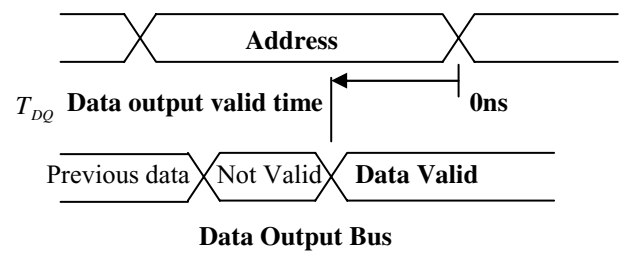

Fig. 7. Timing diagram for data output valid time

In the beginning, we perform the intelligence device characterization learning as described in figure 4, using multiple trip point concepts as described in section 2 and 3. At the end of the learning process (e.g., 50,000 testing patterns applied by ATE, 500,000 training patterns applied by software), a neural network weight file is generated. This file is further used in the sub-optimal worst case test generator. It is called sub-optimal because neural network can not guarantee that the generated output will closely match the perfect approximation. Thus, we further optimize the NN tests using genetic algorithm (GA) as explained in figure 5. At the end of the GA optimization, a set of worst case tests is generated based on WCR. We then further analysed them in detail using industrial ATE and circuit-level simulation. In this paper, we will only discuss the analysis results based on the measurement from industrial ATE. Functional failure detected with our approach is not discussed in this paper.

In order to study the efficiency of our approach, we compare our approach with typical deterministic method and random approach as shown in Table 1. The first column of table 1 is the name of test extracted using different methods as shown in column two. The third column of table 1 shows different WCR values using equation (6)-minimization and the fourth column of table 1 shows different $T_{DQ}$ values extracted from the worst case parameter shmoo in figure 8 based on three different approaches. The shmoo plot shows Vdd power supply in Y-axis, and $T_{DQ}$ timing parameters in X-axis. There are 1000 tests overlapping in a single shmoo plot, so that we can compare the differences between them. Our practical experimental result clearly shows that $T_{DQ}$ is test dependent, as different tests trigger different trip point values in the shmoo plot. This demonstrates the importance of using our multiple trip point concept in device characterization. Moreover, in table 1 the lowest value of $T_{DQ}$ (i.e. the value with highest WCR compared to the other methods) is detected via neural network and genetic algorithm approach.

Table 1 Comparison of $T_{DQ}$ with different approaches: Vdd 1.8V

| Test Name | Technique | WCR | $T_{DQ}$ (ns) |
|---|---|---|---|
| March Test | Deterministic | 0.619 | 32.3ns |
| Random Test | Random | 0.701 | 28.5ns |
| **NNGA Test** | **Neural & Genetic** | **0.904** | **22.1ns** |

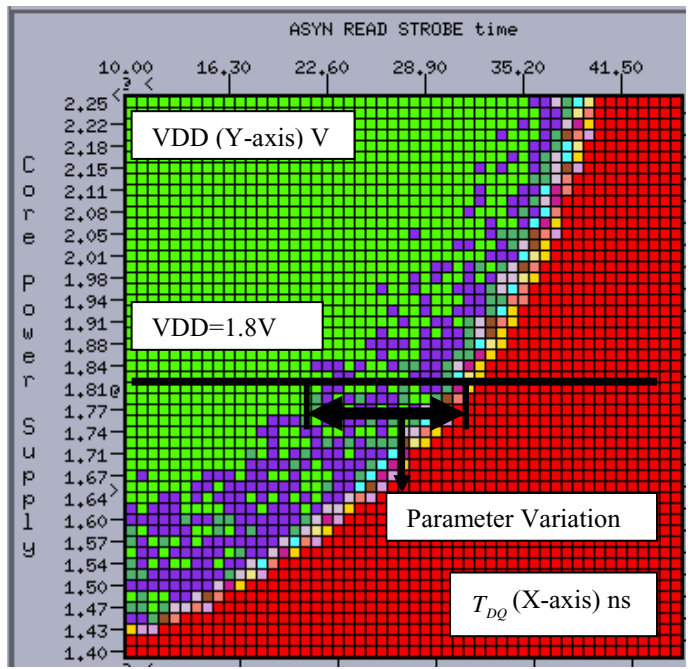

Fig. 8. Shmoo Plot: Worst Case Device Parameter Variation Analysis

## 7. Conclusion

Our experimental results show that the measured device performance is input test dependent. The true worst case test can provoke a large drift of the trip point values, which may lead to degradation of circuit performance or application failure. It is very difficult or not possible at all to obtain this information by any existing conventional single trip point and single test concept. To detect the worst case test, we use computational intelligence techniques with industrial ATE, and we further developed a "find until trip point" method to improve measurement speed and range selection procedure during characterization and search process. To the best of our knowledge, our approach creates the first computational intelligence characterization concept for semiconductor analysis and industrial ATE. This concept changes the focus of typical device characterization. The major benefit of our approach is the fact that it has a high probability to measure the true performance of the device. Although the test time is longer than in a single trip-point method, the cost of our approach will be much lower if we consider the cost of manually performing a detailed analysis on possible circuit-weaknesses left behind by single trip-point method. Thus, our approach is very attractive to get products to market in time.